\documentclass{isprs} 
\usepackage{subfigure}
\usepackage{setspace}
\usepackage{geometry} 
\usepackage{epstopdf}
\usepackage[labelsep=period]{caption}  
\usepackage[british]{babel} 
\usepackage[hang]{footmisc}

\begin{document}

\title{Cooperative Face Liveness Detection from Optical Flow}
\date{}

\author{
 Artem Sokolov\textsuperscript{1, 2}, Mikhail Nikitin\textsuperscript{2}, Anton Konushin\textsuperscript{3, 1}}

\address{
	\textsuperscript{1 }M. V. Lomonosov Moscow State University, Moscow, Russia \\
	\textsuperscript{2 }Tevian, Moscow, Russia -- (artem.sokolov, mikhail.nikitin)@tevian.ru\\
    \textsuperscript{3 }AIRI, Moscow, Russia -- konushin@airi.net\\
}

\abstract{
In this work, we proposed a novel cooperative video-based face liveness detection method based on a new user interaction scenario where participants are instructed to slowly move their frontal-oriented face closer to the camera. This controlled approaching-face protocol, combined with optical flow analysis, represents the core innovation of our approach. By designing a system where users follow this specific movement pattern, we enable robust extraction of facial volume information through neural optical flow estimation, significantly improving discrimination between genuine faces and various presentation attacks (including printed photos, screen displays, masks, and video replays). Our method processes both the predicted optical flows and RGB frames through a neural classifier, effectively leveraging spatial-temporal features for more reliable liveness detection compared to passive methods.
}


\maketitle


\section{Introduction}\label{INTRODUCTION}
 
\par Face recognition (FR) algorithms are used to solve a wide range of practical tasks that require person identification, such as face-payment and crossing point control. Being part of identification systems which are under the threat of dealing with intruders, makes FR algorithms an object for representation attacks. The problem is that FR models treat real and spoof faces, that can be depicted on a piece of paper or a screen, in the same way. Due to this feature, if the identification system is not protected from such attacks, the presence of any person can be easily simulated. To prevent identity substitution during biometric identification, the face liveness detection algorithms are used to determine whether the face in the image is real or not. 

\par The basic face liveness detection methods work with single input image (single-shot), but they are not always able to accurately recognize attacks on the identification systems when high-quality fake faces are used. To increase the accuracy of spoofing classification, the algorithms that take video frames as an input can be utilized (multi-shot). Such an approach gives an opportunity to take into account not only static information about face texture and surrounding, but also temporal information about face movement.
To further improve the accuracy, the input video can be recorded in cooperative mode, when a user is asked to follow a set of instructions while filming. Imposing such restrictions on the process of data collection may increase total time of person identification, but significantly reduces the likelihood of a successful attack on the system.

\par In this work we propose the novel cooperative multi-shot facial liveness detection method. The input videos for the algorithm should be taken following a specific scenario, in which a person is slowly moving frontal-oriented face closer to the camera between two predefined checkpoints ("approaching face" scenario). The optical flow detector is used as the main component of video sequence preprocessing. The novelty of the proposed method lies in the usage of a predicted facial optical flow of the input video, which is filmed by the "approaching face" scenario, for further real/fake classification. The use of optical flow proved to be particularly effective with the chosen video scenario, as this combination gives an opportunity to extract information about face volume. Also, the predicted optical flow is passed to the classifier along with an RGB frame from the video in order to take in account both static and temporal information.

\section{Related Work}\label{RELATED WORK}

\par The field of video-based liveness detection has a long history of research. Some of the early approaches were focused on analyzing specific facial movements, such as eye movement and blinking patterns \cite{eyes1} \cite{eyes3} or lips motion \cite{lips}. While being effective against simple spoofing attempts, these methods proved to be vulnerable to more sophisticated attacks using masks with pre-cut openings in the targeted regions.

\par Another group of video liveness classification approaches is based on adaptation of existing single-shot methods to video scenario. For example, \cite{fqa} introduced a quality assessment module, which gives an opportunity to weight predictions of single-shot model across frames, giving higher importance to predictions on frames, that are easier to classify for single-shot model. However, such methods are not able to fully exploit temporal information as they process frames independently.

\par The emergence of vision transformers enabled new architectures for video analysis. Works like \cite{vitranspad} and \cite{vivit} employed temporal attention mechanisms to capture inter-frame relationships. These approaches provide high classification accuracy, though at significant computatio\-nal cost. 

\par The more efficient approaches are focused on frame aggregation. One of such algorithms was proposed by \cite{parkin2020}. In this method authors combined optical flow detector with rank pooling \cite{rankpool} to extract features from input videos for classification. Another algorithm was proposed by \cite{muhammad2022}, in which stabilized frame averaging was used to compute the sequence aggregation for further liveness classification.

\begin{figure*}[t]
    \includegraphics[width=\textwidth]{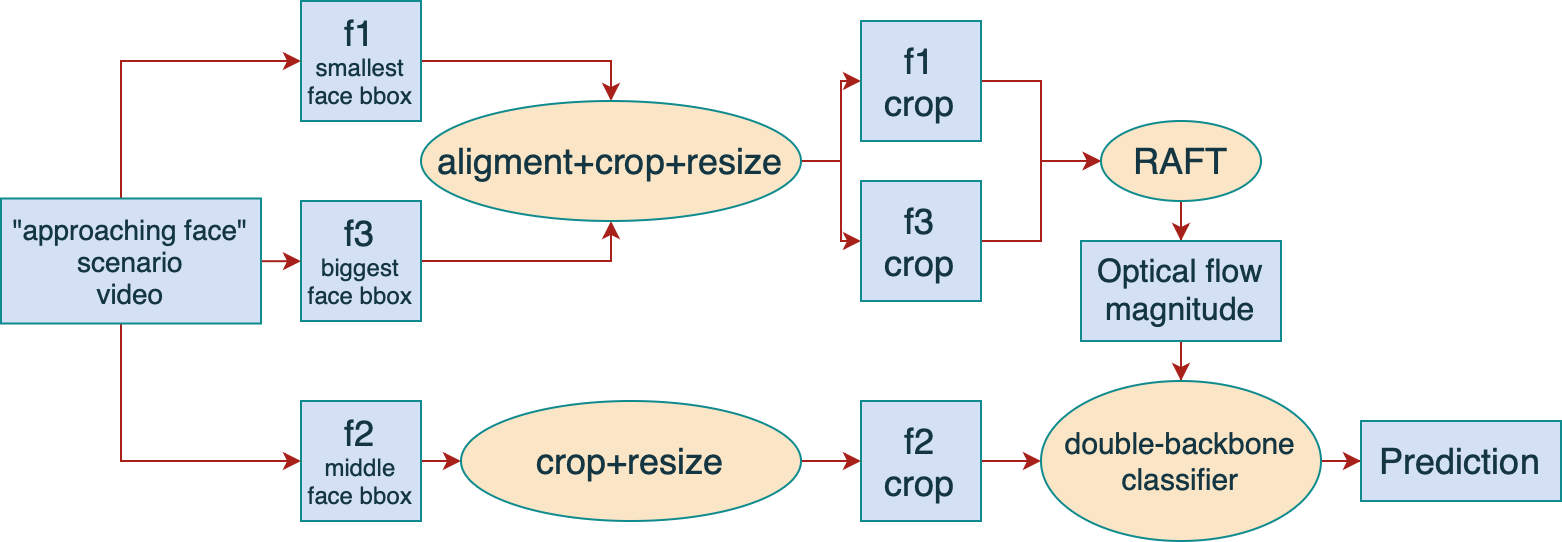}
    \caption{Proposed cooperative liveness detection pipeline}
    \label{pipeline}
\end{figure*}

\par Another approach to the video-based liveness detection is cooperative one, which remains relatively underexplored in literature, with most methods relying on simple interactions, like sentence pronunciation \cite{lips} or facial expression changes \cite{ming2018facelivenet}. Such approaches increase the time of verification, but significantly improve the accuracy.

\par Our method builds upon these foundations while introducing key innovations. Like \cite{muhammad2022}, we employ stabilization, but enhance it with neural key-point detection for improved alignment. Similar to \cite{parkin2020}, we utilize optical flow, but specifically optimize it for our uni\-que cooperative scenario where controlled face movement pro\-vides crucial volumetric information unavailable in passive ap\-proaches. Furthermore, we combine optical flow analysis with RGB frame processing to capture both dynamic and static features, improving robustness across diverse attack types from simple printouts to sophisticated 3D masks.

\section{Proposed Method}\label{PROPOSED METHOD}

\par Our pipeline (Figure \ref{pipeline}) collects and processes cooperative vi\-deos through five stages:
\begin{enumerate}
    \item \textbf{Input frames capturing}: The first (f1), the middle (f2), and the last (f3) frames are extracted according to the cooperative scenario (details in Section \ref{Cooperative scenario});
    \item \textbf{Frame preprocessing}: The frames f1, f2, and f3 are alig\-ned and normalized (details in Section \ref{preproc}).
    \item \textbf{Optical flow computation}: We calculate optical flow bet\-ween the f1 and the f3 using the pretrained RAFT detector \cite{teed2020raft}.
    \item \textbf{Optical flow processing}: The magnitude is calculated by the predicted optical flow and then clipped (details in Section \ref{OF}).
    \item \textbf{Classification}: The processed optical flow and the RGB frame f2 are fed to the neural classifier for final prediction (details in Section \ref{sub:cls_model}).
\end{enumerate}

\subsection{Cooperative scenario}\label{Cooperative scenario}

\begin{figure}[h]
    \centering
    \includegraphics[width=0.25\textwidth]{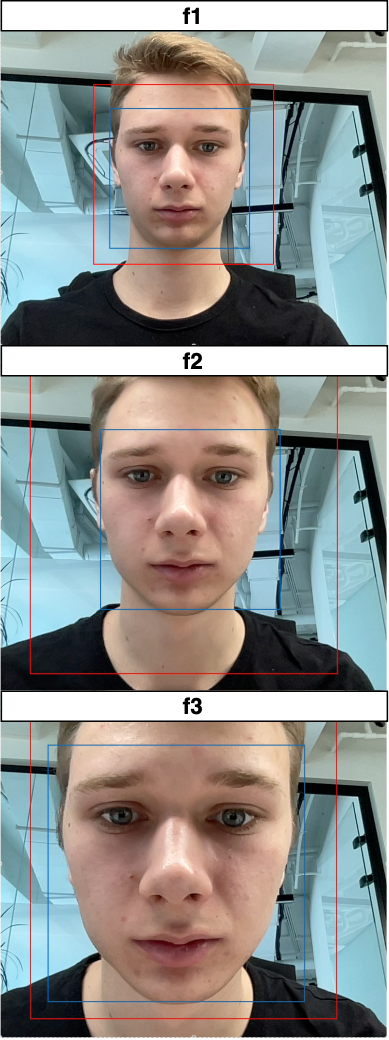}
    \caption{Proposed cooperative scenario}
    \label{scenario}
\end{figure}

The system guides users through the standardized recording protocol visualizing recorded video on the display  (Figure \ref{scenario}). A real-time face detector displays the current face bounding box (blue square). The user first has to align the face with a red reference square (50\% frame height), then slowly move it closer to the camera until the face fills an enlarged target square (75\% frame height). If the face moves back from the camera or disappears, the recording starts from the beginning. This yields three key frames (f1, f2, f3) with relative face heights of 0.500, 0.625, and 0.750 respectively, that are extracted from the recorded video.

\subsection{Input preprocessing}\label{preproc}
\par
\begin{figure}[h]
    \centering
    \includegraphics[width=0.45\textwidth]{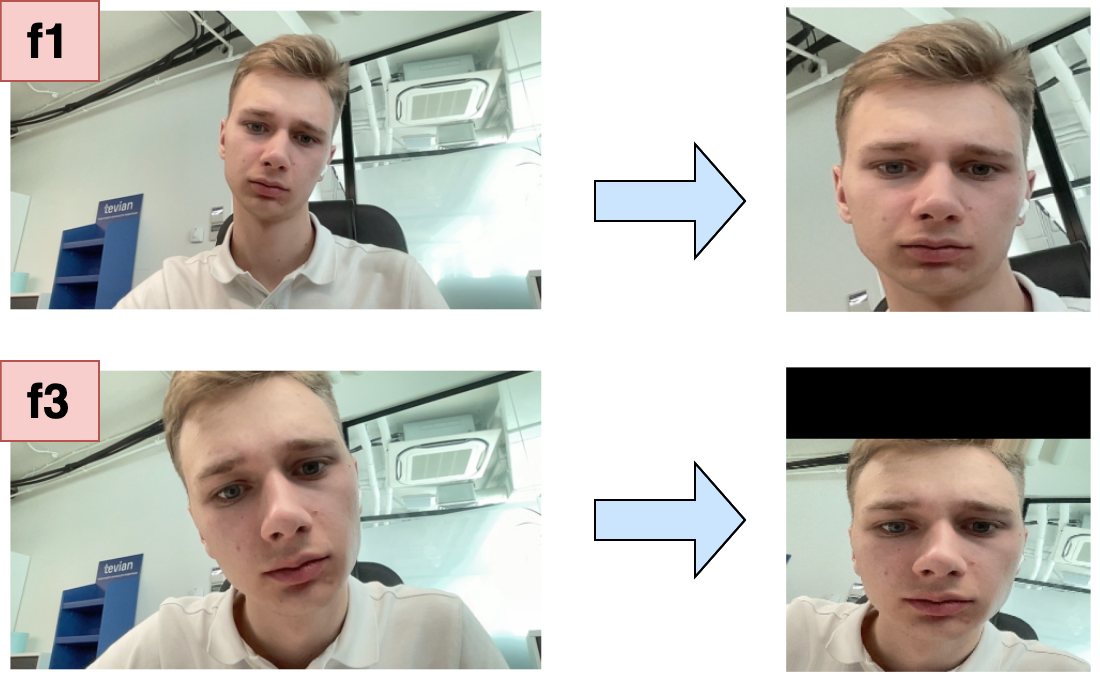}
    \caption{Preprocessing of the f1 and the f3 frames before optical flow computation}
    \label{preprocessing}
\end{figure}

 In order to make optical flow detector able to precisely extract facial movements between the first and the last frames of the video with significant face size variation, we apply specialized preprocessing (Figure \ref{preprocessing}):

\begin{enumerate}
    \item Detect facial key-points using a neural detector;
    \item Transform f1 and f3 frames through shift and rotation operations to center and align their key-points;
    \item Crop the face region with 10\% margin around the bounding box;
    \item Resize all crops to 256×256 pixels.
\end{enumerate}

\par The f2 frame undergoes identical processing except for the \\ alignment step.

\subsection{Optical flow}\label{OF}

\par
\begin{figure}[h!]
    \centering
    \subfigure[]{\includegraphics[width=0.22\textwidth]{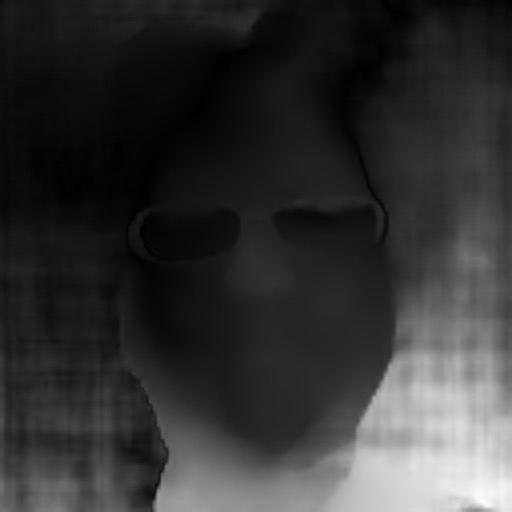}} 
    \subfigure[]{\includegraphics[width=0.22\textwidth]{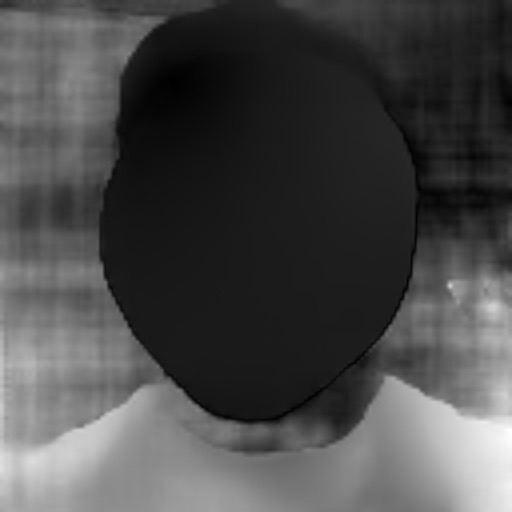}} 
    \subfigure[]{\includegraphics[width=0.22\textwidth]{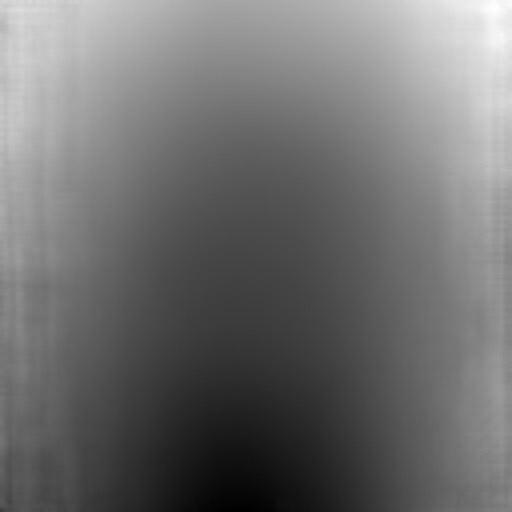}}
    \subfigure[]{\includegraphics[width=0.22\textwidth]{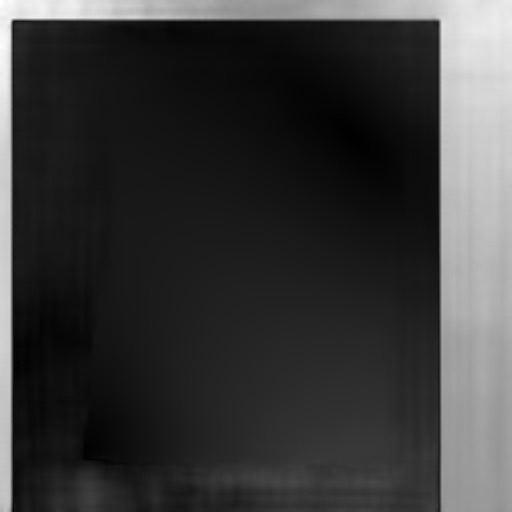}}
    \caption{Optical flow magnitude examples: (a) Real face (b) Printed mask attack (c) Screen photo attack (d) Printed photo attack}
    \label{fig:flow_examples}
\end{figure}

 We compute optical flow between preprocessed f1 and f3 frames using pretrained RAFT \cite{teed2020raft}. This reveals distinct patterns for different attack types (Figure \ref{fig:flow_examples}):
\par \textbf{Flat static spoofs} (print photos / screen photos).  Such fakes' magnitudes show near-zero flow magnitude value in square region around the face (Figure \ref{fig:flow_examples}c-d).
\par \textbf{Flat masks.} Such fakes' magnitudes show close to zero flow magnitudes in a face region and high values on the background (Figure \ref{fig:flow_examples}, (b)).
\par \textbf{Real faces and dynamic video replays}. Optical flow magnitudes are similar to masks' ones, but demonstrates 3D face patterns and more smooth edges between head and background (Figure \ref{fig:flow_examples}, (a)).

\par Since background information is non-informative, we clip magnitude values exceeding 20\% of the crop size (51.2 pixels for 256×256 inputs).

\subsection{Classification model}\label{sub:cls_model}

\par
\begin{figure}[h]
    \centering
    \includegraphics[width=0.4\textwidth]{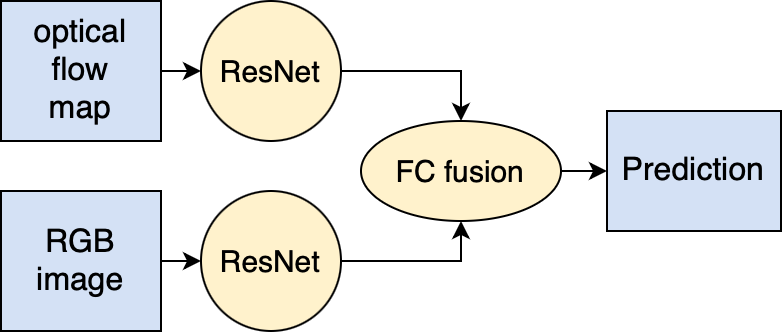}
    \caption{Double-backbone architecture for liveness classification by optical flow map and RGB image}
    \label{Architecture}
\end{figure}

 For evaluation, we employ a dual-backbone ResNet18 architecture with fully connected fusion layer (Figure \ref{Architecture}). The first "flow backbone" processes the clipped optical flow magnitude between f1 and f3. The second "RGB backbone" processes the preprocessed f2 frame.

\subsection{Training details}\label{sub:training}

\par For proposed method evaluation both backbones are trained simultaneously. Though in practice, the amount of available data for single shot liveness classification is disproportionately larger than can be collected for the chosen "approaching face" scenario. That is why RGB backbone can be separately trained on larger single-shot datasets.

\par In order to expand the training set we employ several optical flow-specific augmentations:

\begin{itemize}
    \item \textbf{Random frame}: Randomly select f1 from the top 10\% frames with smallest faces and f3 from the top 10\% frames with largest faces;
    \item \textbf{Multi resolution}: Compute flows at resolutions from \\192x192 to 320x320 and then resize to 256×256;
    \item \textbf{Perspective augmentation}: Apply random perspective\\ transforms to the f1 and f3 frames before preprocessing to improve robustness to head rotations.
\end{itemize}

\section{Data and Experiments}\label{DATA AND EXPERIMENTS}
\subsection{Dataset}\label{Dataset}

\par Since our proposed method relies on a novel "approaching face" capture scenario, existing public datasets are unsuitable for training and evaluation, as they do not adhere to this specific protocol. To address this gap, we collected a dedicated private dataset. All samples were captured using diverse devices and lighting conditions, with recordings contributed by multiple participants. The dataset strictly follows the "approaching face" scenario, containing videos of both genuine faces and various types of spoofs (e.g., printed photos, screen replays). Detailed sample statistics are provided in Table \ref{tab:Private_statistics}. The dataset comprises the following subsets:

\begin{itemize}
    \item \textbf{Real.} Set of images with real human faces;
    \item \textbf{Screen Photos.} The subsample shot with photographs
          of faces demonstrated on various displays;
    \item \textbf{Printed Photos.} The subsample shot with printed photographs;
    \item \textbf{Printed Masks.} The subsample shot with printed full-face masks and with cut-out facial regions;
    \item \textbf{Dynamic Videos.} The subsample shot with prerecorded videos, filmed according to "approaching face" scenario;
    \item \textbf{Static Videos.} The subsample shot with prerecorded vide\-os with almost static faces;
\end{itemize}

\begin{table}[h!]
	\centering
        \renewcommand{\arraystretch}{1.2}%
		\begin{tabular}{|l|c|c|}
        \hline
			 Videos type & train split & test split \\
        \hline
			 Real&1860&291\\
			 Screen Photos &4623&217\\
			 Printed Photos&1679&85\\
			 Printed Masks&1699&93\\
			 Dynamic Videos&870&170\\
			 Static Videos&765&218\\\hline
		\end{tabular}
	\caption{Private dataset statistics (number of samples).}
\label{tab:Private_statistics}
\end{table}

\subsection{Evaluation metrics}

\par We measured performance using the ROC AUC metric, computed separately for each spoofing subset (class 0) against the real-face subset (class 1).

\subsection{RAFT parameters selection}

\par
\begin{center}
        \begin{figure}[h]
            \centering
            \includegraphics[width=1.0\linewidth]{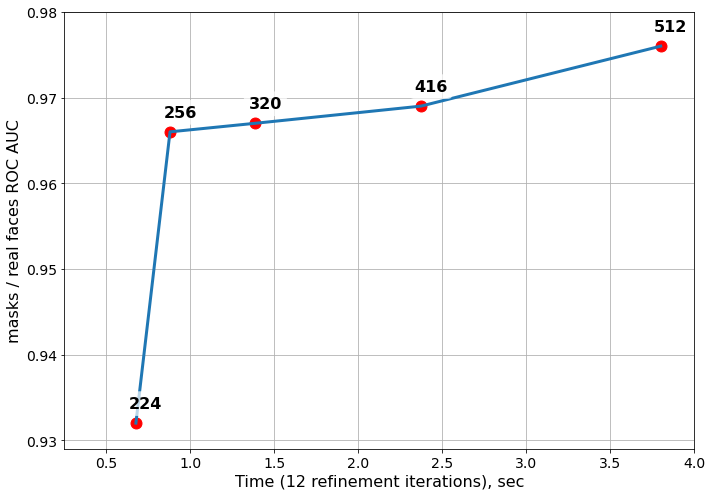}
            \caption{Selection of the OF resolution}
            \label{fig:resolution}
        \end{figure}
\end{center}

\begin{center}
        \begin{figure}[h]
            \centering
            \includegraphics[width=1.0\linewidth]{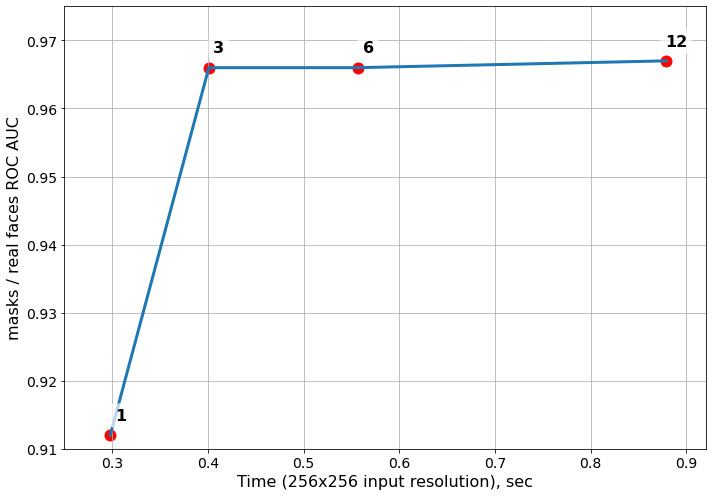}
            \caption{Selection of the RAFT refinement iterations number}
            \label{fig:iters}
        \end{figure}
\end{center}

 The optical flow detector’s resolution and number of refinement iterations significantly impact both computational efficiency \\and face volume information extraction. As the optical flow computation is the bottleneck of the proposed pipeline, we tried to speed it up by using the input data of the smaller resolution and by shortening the number of refinement iterations. For further experiments 256x256 resolution and 3 refinement iterations were fixed, according to experiments shown on Figures \ref{fig:resolution} and \ref{fig:iters}.

\par All the time measurements were conducted in a single thread mode on i5-11600K CPU.

\subsection{Ablation study}

\par As the baseline model we use the single-ResNet18 classifier that takes optical flow map of size 256x256 as an input. In all our experiments, optical flow map is computed using 3 refinement iterations of RAFT.

\subsubsection{Optical flow processing}
Several options of optical \\flow processing were considered:
\begin{itemize}
    \item \textbf{Raw optical flows}. Two-channel input with X/Y-axis pix\-el displacements;
    \item \textbf{Optical flow magnitudes}. Single-channel input with displacement distances for each pixel:\\$magnitude_{ij} = \sqrt{(x\_shift_{ij} - y\_shift_{ij})^2}$, where $i,j$ the coordinates of a pixel;
    \item \textbf{Clipped optical flow magnitudes}. The magnitude value which are greater than 20\% of the input crop side are clip\-ped to $inp\_size* 0.2$.
\end{itemize}

\par The table \ref{tab:OF proc} shows that the ROC AUC of the Dynamic Videos subsample is significantly lower than the others. This is due to the fact that the such spoofs' optical flows are similar with the real faces' ones.
\par In further experiments we use cropped optical flow magnitudes as the input of the classifier.

    \begin{table*}[h!]
	\centering
    \renewcommand{\arraystretch}{1.1}%
		\begin{tabular}{|l|c|c|c|c|c|}\hline
             OF preprocessing type&Screen&Printed&Masks&Dynamic Videos&Static Videos \\\hline
             OF &0.992&0.990&0.953&0.668&0.958\\
			 OF magnitude&0.991&0.991&0.955&0.670&0.961\\
			 clipped OF magnitude&\textbf{0.994}&\textbf{0.993}&\textbf{0.965}&\textbf{0.784}&\textbf{0.965}\\\hline
    		\end{tabular}
    	\caption{Comparison of optical flow preprocessing (ROC AUC)}
    \label{tab:OF proc}
    \end{table*}

\begin{table*}[h!]
	\centering
    \renewcommand{\arraystretch}{1.1}%
		\begin{tabular}{|l|c|c|c|c|c|}\hline
             Augmentations&Screen&Printed&Masks&Dynamic Videos&Static Videos \\\hline
             no augmentations &0.994&0.993&0.965&0.784&0.965\\
			 random frame&\textbf{0.998}&\textbf{0.997}&0.972&0.666&0.972\\
			 random frame + multi resolution&\textbf{0.998}&0.996&\textbf{0.972}&\textbf{0.675}&\textbf{0.972}\\\hline
    		\end{tabular}
    	\caption{Comparison of optical flow augmentations (ROC AUC)}
    \label{tab:OF aug}
    \end{table*}

\begin{table*}[h!]
\centering
\renewcommand{\arraystretch}{1.1}%
    \begin{tabular}{|l|c|c|c|c|c|}\hline
         Architecture&Screen&Printed&Masks&Dynamic Videos&Static Videos \\\hline
         OF only; ResNet18 &0.998&0.996&0.972&0.675&0.972\\
         OF + RGB (stacked); single ResNet18&0.996&0.994&0.985&0.991&0.998\\
         OF + RGB (separately); double ResNet18&\textbf{0.999}&\textbf{1.000}&\textbf{0.993}&\textbf{0.994}&\textbf{0.999}\\\hline
        \end{tabular}
    \caption{Comparison of architectures (ROC AUC)}
\label{tab:Architecture}
\end{table*}

\begin{table*}[h!]
	\centering
    \renewcommand{\arraystretch}{1.1}%
		\begin{tabular}{|l|c|c|c|c|c|}\hline
             Modification&Screen&Printed&Masks&Dynamic Videos&Static Videos \\\hline
             None &0.999&1.000&0.993&0.994&0.999\\
			 perspective augmentations&\textbf{1.000}&\textbf{1.000}&\textbf{0.994}&\textbf{0.996}&\textbf{0.999}\\
			 perspective augmentations + stabilization &\textbf{1.000}&\textbf{1.000}&\textbf{0.994}&\textbf{0.996}&\textbf{0.999}\\\hline
    		\end{tabular}
    	\caption{Modifications for increacing model stability on samples with head rotations  (ROC AUC)}
    \label{tab:rotate}
    \end{table*}

\begin{table*}[h!]
	\centering
    \renewcommand{\arraystretch}{1.1}%
		\begin{tabular}{|l|c|c|c|c|c|}\hline
             Blurriness level&Screen&Printed&Masks&Dynamic Videos&Static Videos \\\hline
             no blur &1.000&1.000&0.994&0.996&0.999\\
             low &0.997&0.996&0.992&0.989&0.999\\
             medium &0.990&0.981&0.939&0.966&0.984\\
			 high &0.937&0.883&0.833&0.846&0.857\\\hline
    		\end{tabular}
    	\caption{Evaluation on the blurred test (ROC AUC)}
    \label{tab:blur}
\end{table*}

\begin{table*}[h!]
	\centering
    \renewcommand{\arraystretch}{1.1}%
		\begin{tabular}{|l|c|c|c|c|c|}\hline
             Aggregation&Screen&Printed&Masks&Dynamic Videos&Static Videos \\\hline
             single shot &0.983&0.963&0.970&0.991&0.985\\
             stabilized average \cite{muhammad2022} &0.999&0.980&0.922&\textbf{1.000}&\textbf{1.000}\\
             rank pool + OF \cite{parkin2020} &0.993&0.982&0.904&0.921&0.976\\
			 proposed method &\textbf{1.000}&\textbf{1.000}&\textbf{0.994}&0.996&0.999\\\hline
    		\end{tabular}
    	\caption{Aggregation types comparison (ROC AUC)}
    \label{tab:aggreg}
    \end{table*}


\subsubsection{Optical flow augmentations}

\par To extend train set of optical flows we use augmentations, discussed in \ref{sub:training}.
\par The table \ref{tab:OF aug} shows that the \textbf{random frame} augmentation promoted significant ROC AUC growth, while the \textbf{multi resolution} did not affected metrics. However, for overall classifier stability we use both augmentations in further experiments and final pipeline.

\subsubsection{RGB input}

\par In order to make the model extract textural information from input frames, we modified the architecture to make it able to receive an RGB image along with an optical flow. These are two proposed architecture variants that we evaluated:
\begin{itemize}
    \item \textbf{OF + RGB (stacked); single ResNet18}. Input is a concatenation of the RGB image and the optical flow magnitude;
    \item \textbf{OF + RGB (separately); double ResNet18}. The first backbone processes an optical flow magnitude, the second one processes an RGB image.
\end{itemize}

\par The experiments, described in the table \ref{tab:Architecture} shows that the model ability to extract textural information significantly increases ROC AUC metric for all subsets, especially for the Dynamic Videos.

\par In the further experiments we use the dual-ResNet18 architecture as it outperformed other models on all data subsets.

\subsubsection{Head rotations}
\par To increase the model stability to \\head rotations between the frames f1 and f3 the following measures have been taken:
\begin{itemize}
    \item \textbf{Face stabilization by facial key-points}. This step in input preprocessing makes model stable to roll rotations;
    \item \textbf{Random perspective transformations}. This augmentation is applied to f1 and f3 images before preprocessing, making model stable to pitch and yaw rotations.
\end{itemize}

\par The table \ref{tab:rotate} shows that the ROC AUC metric does not affected by the discussed changes. That happened, because the current test split does not contain samples with strong head rotations. However, manual tests shows that the pipeline becomes stable to all types of head rotations after adding proposed modifications to the final pipeline.

\subsubsection{Blur}

\par We checked the stability of the proposed pipeline on blurred images. We used Gaussian Blur function from \\OpenCV to create images with blur. In our experiments we used three levels of blurriness: low (kernel size is 1\% of the mean f1 face crops hight), medium (kernel size is 6\% of the mean f1 face crops hight) and high (kernel size is 12\% of the mean f1 face crops hight) (Table \ref{tab:blur}).

\subsection{Comparison with other aggregation methods}

\par We also adapted ideas from \cite{muhammad2022} and \cite{parkin2020} to the proposed cooperative scenario to make a comparison with developed pipeline (Tables \ref{tab:aggreg}, \ref{tab:time}). These methods were chosen for the comparison as we found them profitable to use with "approaching face" scenario. Here are the details of the methods adaptation:

\subsubsection{Classification by stabilized average frame}
\par The idea was taken from \cite{muhammad2022}. In order to improve face stabilization we use the neural facial key-points detector. The conducted experiments showed that the best algorithm quality is achieved with using 5 frames for computing input video aggregation. We used ResNet18 model for classification.

\subsubsection{Classification by optical flow and rank pooling outputs}
\par The idea was taken from \cite{parkin2020}. In the adapted pipeline we apply rank pooling to the input vide\-os twice with the same parameters as in the original paper. Also we compute optical flow between the f1 and the f3 frames with RAFT as in the proposed method. The received features are passed to three ResNet18 backbones with fully connected fusion layer as in the proposed method.

\par Also we compared proposed pipelines with the single-shot ResNet18 model, that was evaluated on the f2 frames.

\begin{table}[h!]
	\centering
    \renewcommand{\arraystretch}{1.1}%
		\begin{tabular}{|l|c|c|c|c|c|}\hline
             Method& sec. \\\hline
             single-shot & 0.05\\
             stabilized average & 0.05\\
             rank pool + OF & 0.75\\
			 proposed method & 0.55\\\hline
    		\end{tabular}
    	\caption{Runtime performance comparison (without preprocessing)}
    \label{tab:time}
    \end{table}

\section{Conclusion}\label{CONCLUSION}

\par In this work, we proposed the novel cooperative video-based face liveness detection method that leverages optical flow analysis under a controlled "approaching face" scenario. Our approach combines stabilized facial key-point alignment with \\ neural optical flow estimation to effectively distinguish between real faces and various types of spoofing attacks, including printed photos, screen displays, masks, and video replays.

{
	\begin{spacing}{1.17}
		\normalsize
		\bibliography{ISPRSguidelines_authors} 
	\end{spacing}
}

\end{document}